\documentclass[letterpaper, 10 pt, conference]{ieeeconf}
\IEEEoverridecommandlockouts
\usepackage{cite}
\usepackage{amsmath,amssymb,amsfonts}
\usepackage[linesnumbered, ruled]{algorithm2e}
\usepackage[flushleft]{threeparttable}
\usepackage{siunitx} 
\usepackage{graphicx}
\usepackage{textcomp}
\usepackage{algpseudocode}
\usepackage{float}
\usepackage{subfig}
\usepackage{xcolor}
\usepackage{booktabs}
\usepackage{overpic}  
\usepackage[export]{adjustbox}  
\usepackage{multirow}  
\usepackage{bm}  
\usepackage{cellspace}  
\usepackage{wrapfig}  
\usepackage{todonotes}

\def\BibTeX{{\rm B\kern-.05em{\sc i\kern-.025em b}\kern-.08em
    T\kern-.1667em\lower.7ex\hbox{E}\kern-.125emX}}

\title{VISO-Grasp: Vision-Language Informed Spatial Object-centric 6-DoF Active View Planning and Grasping in Clutter and Invisibility\\
\thanks{Karlsruhe Institute of Technology, Karlsruhe, Germany. Email: yitian.shi@kit.edu. This work is sponsored by the DFG SFB-1574 Circular Factory project.} 
\thanks{*Equal contribution}

}

\author{Yitian Shi*, 
        Di Wen*,
        Guanqi Chen,
        Edgar Welte,
	Sheng Liu,
        Kunyu Peng, 
        Rainer Stiefelhagen,
        Rania Rayyes
 }

\begin{document}
\maketitle
\thispagestyle{empty}
\pagestyle{empty}

\begin{abstract}
We propose VISO-Grasp, a novel vision-language-informed system designed to systematically address visibility constraints for grasping in severely occluded environments. By leveraging Foundation Models (FMs) for spatial reasoning and active view planning, our framework constructs and updates an instance-centric representation of spatial relationships, enhancing grasp success under challenging occlusions. Furthermore, this representation facilitates active Next-Best-View (NBV) planning and optimizes sequential grasping strategies when direct grasping is infeasible. Additionally, we introduce a multi-view uncertainty-driven grasp fusion mechanism that refines grasp confidence and directional uncertainty in real-time, ensuring robust and stable grasp execution. Extensive real-world experiments demonstrate that VISO-Grasp achieves a success rate of $87.5\%$ in target-oriented grasping with the fewest grasp attempts outperforming baselines. To the best of our knowledge, VISO-Grasp is the first unified framework integrating FMs into target-aware active view planning and 6-DoF grasping in environments with severe occlusions and entire invisibility constraints. Code is available at: https://github.com/YitianShi/vMF-Contact

\end{abstract}

\section{Introduction}

Robotic grasping in unstructured, cluttered environments remains a significant challenge, particularly for executing target-oriented grasps under partial or complete occlusions~\cite{newbury2023deep}.
Humans instinctively overcome occlusion during targeted searches by adjusting their viewpoints and intuitive reasoning about spatial relationships to infer potential target locations. In comparison, leveraging the recent success of deep neural networks and contributions to large-scale training data~\cite{fang2020graspnet, gilles2023metagraspnetv2, mahler2017dex, murali20206}, existing learning-based robotic grasp detectors~\cite{newbury2023deep} rely primarily on static viewpoint observations, assuming sufficient visibility of the target object from single or pre-defined multi-view perspectives\cite{dai2023graspnerf, breyer2021volumetric, 10892643}. This constraint reduces flexibility in scenarios where the target object is highly occluded or entirely unobservable. 

Meanwhile, recent active grasping frameworks~\cite{breyer2022closed, zhang2023affordance, NEURIPS2024_4364fef0} have made initial attempts to integrate Next-Best-View (NBV) planning into grasping, which prioritize 3D reconstruction and rendering over target-driven view search for grasping, relying heavily on pre-defined search spaces. 
However, without such prior knowledge, these methods may fail to handle heavily cluttered scenarios and target invisibility. We argue that effectively reaching the target, even under invisibility, requires a structured active vision and sequential grasping strategy. This shall leverage object relationships to systematically optimize sensor viewpoints, remove obstructions when necessary, and incrementally reveal the target for efficient and successful grasp execution.

To this end, we propose VISO-Grasp, a Vision-language Informed Spatial Object-centric grasping framework that integrates off-the-shelf Foundation Models (FMs) \cite{bai2025qwen2, ravi2024sam2, liu2024grounding} with active vision and occlusion-aware $\mathrm{SE(3)}$ grasp planning, which leverages the inherent prior of VLMs to achieve human-like decision-making. 
In contrast to recent works~\cite{noh2024graspsam, tziafas2024openworldgraspinglargevisionlanguage, xu2024rt, dengler2025efficient}, which focus on planar 4-DoF grasp reasoning with VLMs while overlooking their fundamental limitations in accurate 3D spatial understanding and real-time execution, VISO-Grasp introduces evolving spatial reasoning that continuously integrates spatial and occlusion relationships. By incorporating an online grasp fusion mechanism, our approach dynamically refines target visibility and substantially improves grasp efficiency in heavy occlusions. Specifically, when direct grasping is infeasible, a sequential grasp strategy guided by VLMs is employed, leveraging the observable relational hierarchy to determine the optimal grasp sequence for decluttering and removing potential obstructions. In summary, our main contributions are threefold:

\begin{figure}
    \centering
    \includegraphics[width=.96\linewidth]{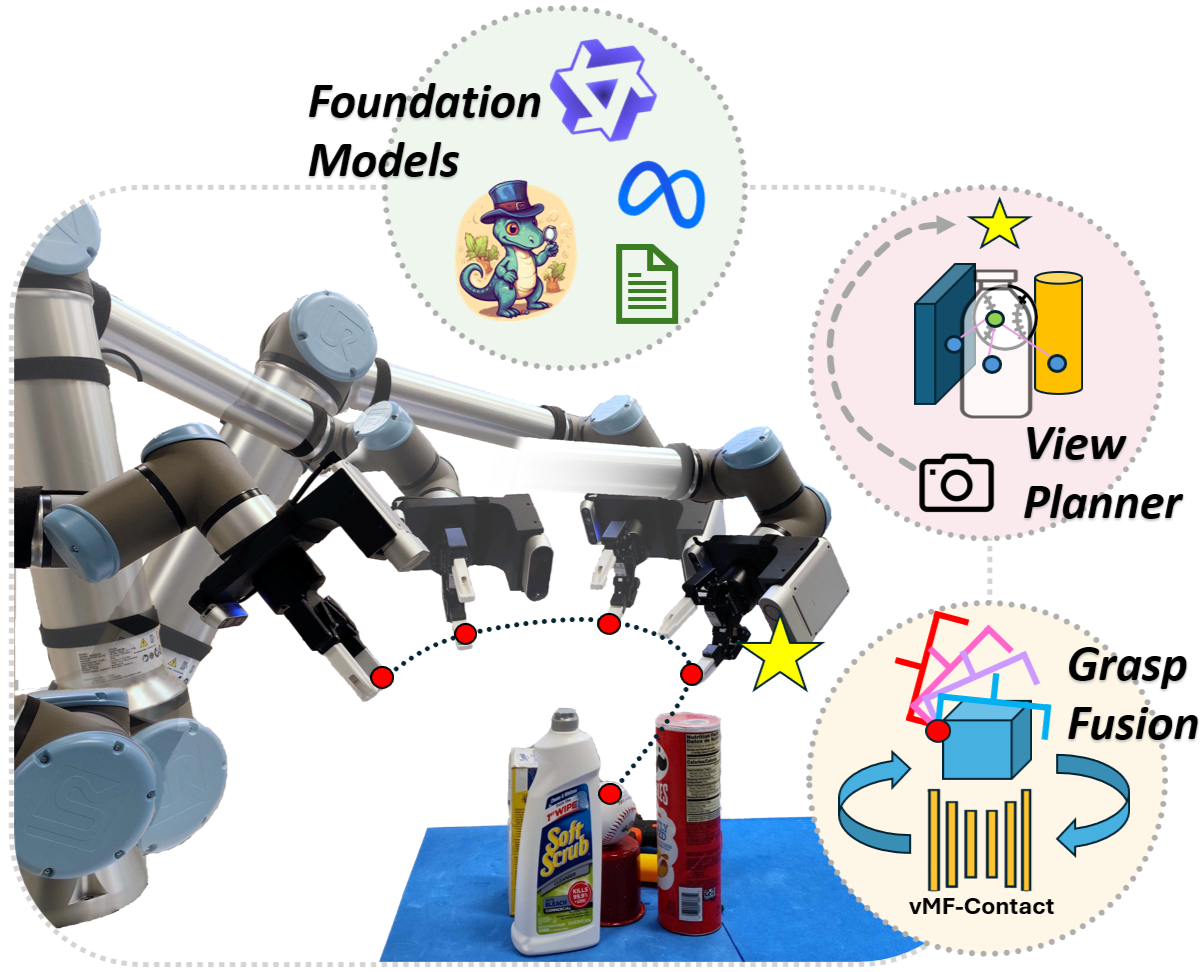} 
    \caption{VISO-Grasp, a unified system integrating Foundation Models (FMs) into target-aware active view planning and uncertainty-driven real-time 6-DoF grasp fusion.
    }
    \label{fig:highlight}
\end{figure}

- \textbf{Vision-language-driven unified scene understanding for robotic perception}, which integrates scene reasoning from the VLM~\cite{bai2025qwen2} to infer spatial relationships between objects. This facilitates occluder identification through Visual-In-Context Learning (VICL), and determines optimal sequential grasp orders in highly occluded environments. 

- \textbf{Target-aware view planning for object-centric grasping}, a novel NBV planner that constructs object-centric continuous velocity fields using 3D instance segmentations from existing foundation models~\cite{liu2024grounding, ravi2024sam2} to optimize sensor viewpoints and maximize target visibility in cluttered environments.

- \textbf{Uncertainty-driven online grasp fusion}, a real-time grasp fusion framework that integrates multi-view grasp predictions through pose-centric categorization and real-time Bayesian updates. This leverages a pre-trained uncertainty-aware grasp generator~\cite{shi2024vmf}, exclusively on simulated data, to enhance grasp robustness and reliability.

\begin{figure*}[h]
    \centering    \includegraphics[width=.9\textwidth]{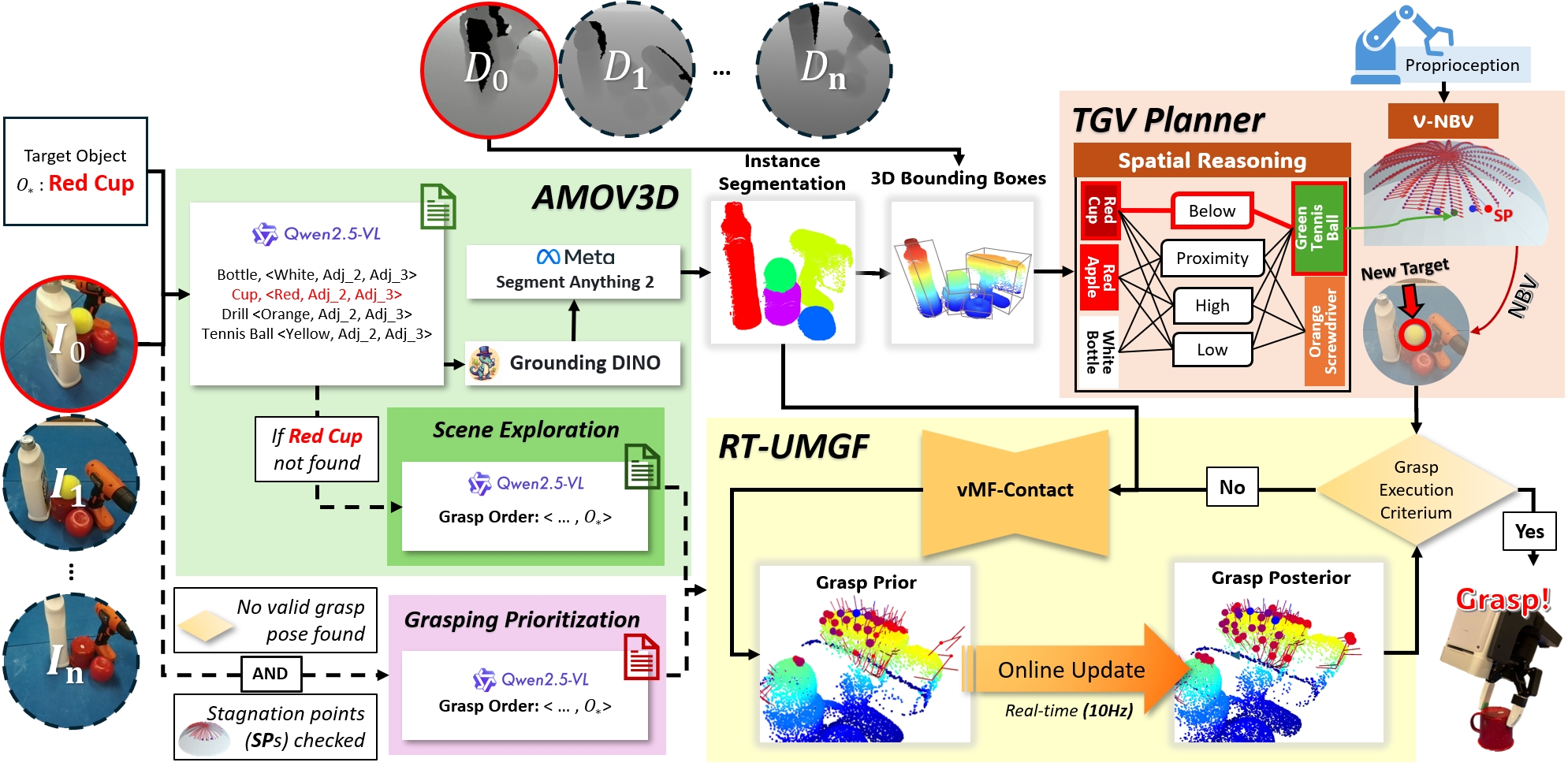}
    \caption{Overview of VISO-Grasp. Our system consists of: 1) Adaptive Multi-View Open-Vocabulary 3D Object Detector (AMOV3D, Sec. \ref{sec:AMOV3D}): It leverages VLM (Qwen2.5-VL~\cite{bai2025qwen2}), Grounding DINO~\cite{liu2024grounding}  and Segment Anything 2~\cite{ravi2024sam2} to extract open-vocabulary 3D object representations. 2) Target-Guided View Planning (TGV-Planner, Sec.~\ref{sec:TGV-Planner}): The TGV-Planner dynamically adjusts viewpoints using spatial reasoning and \textit{Velocity-field-based NBV (V-NBV)} to enhance object discovery. 
    3) Real-Time Uncertainty-guided Multi-view Grasp Fusion (RT-UMGF, Sec.~\ref{sec:RT-UMGF}) to enable occlusion-aware object-centric grasping, RT-UMGF fuses inferred grasps through real-time uncertainty-aware updates in 10Hz.}
    \vspace{-.4cm}
\label{fig:viso}
\end{figure*}
\section{Related works}
\subsection{Active View Search for Grasp Detection}
Recent advancements in scene-level $\mathrm{SE(3)}$ grasp detection frameworks prioritize generalization to unseen, unstructured environments with diverse objects. While end-to-end approaches~\cite{sundermeyer2021contact, Liang_2019, breyer2021volumetric, jiang2021synergies} employ deep neural networks to infer grasp configurations from point clouds~\cite{sundermeyer2021contact, Liang_2019} or volumetric representations ~\cite{breyer2021volumetric, jiang2021synergies}, the robustness diminishes in the presence of sim-to-real gap and heavy occlusions, where incomplete geometric information significantly impairs prediction reliability. To this end, multi-view strategies~\cite{morrison2019multi, dai2023graspnerf, 10892643, breyer2021volumetric} integrate observations across multiple viewpoints, while, either bringing the cost of target-agnostic view exploration or relying on pre-defined perspectives by sacrificing flexibility. 

Within this context, active vision~\cite{bajcsy2018revisiting} facilitated by NBV strategies, emerges as a principled approach in vision-based decision-making such as navigation~\cite{gonzalez2002navigation, jin2024activegs} and 3D reconstruction~\cite{pan2024many, zeng2020pc}. By incorporating historical observations, active vision employs a closed-loop optimization process to iteratively select viewpoints and maximize scene visibility for downstream tasks. Recent active grasping frameworks~\cite{breyer2022closed, zhang2023affordance, NEURIPS2024_4364fef0} primarily formulate the NBV problem as a correlated reconstruction and rendering task. For instance, they leverage strategies such as maximization of volumetric Information Gain (IG) derived from ray casting~\cite{breyer2022closed} and minimization of neural graspness or rendering inconsistency~\cite{NEURIPS2024_4364fef0, zhang2023affordance}. These approaches are constrained by discrete viewpoint sampling, pre-defined search space prior (e.g., the bounding box around the targets), target-agnostic grasp planning, and the prioritization of reconstruction or rendering over grasping. Consequently, these limitations hinder adaptability for optimal target-oriented grasp and view planning in unstructured environments or even invisibility. In comparison, we consider involving spatial reasoning between objects to realize object-centric active view planning. 

\subsection{FMs in Robotic Perception and Planning}

    Foundation Models (FMs), including VLMs, have significantly improved robotic perception and planning by zero-shot object detection~\cite{liu2023partslip}, multi-modal reasoning~\cite{sharma2021skill,xu2024rt,jin2024reasoning}, and decision-making~\cite{sun2023smart,radosavovic2023real}, reducing reliance on task-specific data collection while enhancing adaptability in real-world environments~\cite{ firoozi2023foundation, hu2023toward}. Recent works leverage FMs to enable open-vocabulary perception, integrating zero-shot segmentation for novel object recognition ~\cite{tziafas2024openworldgraspinglargevisionlanguage,noh2024graspsam}, while uncertainty-aware frameworks refine scene understanding by dynamically calibrating perception and decision uncertainties~\cite{bhatt2024know}. In robotic planning, multimodal models facilitate goal-conditioned decision-making by aligning spatial and semantic representations~\cite{tang2025foundationgrasp}, while vision-language systems enhance strategic scene exploration and adaptive reasoning in unstructured scenarios~\cite{qianthinkgrasp}. Yet, most FM-driven approaches still rely on 2D representations, lacking explicit 3D spatial priors and depth reasoning~\cite{sripada2024ap}. Furthermore, they assume minimal occlusions and primarily operate under SE(2) planar constraints, with limited real-time viewpoint adaptation and active perception to dynamically refine scene understanding. To overcome these limitations, we propose the first FMs-driven framework that unifies multi-view 3D spatial reasoning and active viewpoint adaptation for occlusion-aware 6-DoF grasping, enabling dynamic scene refinement and grasp execution in cluttered environments.


\section{Problem Formulation}
\label{sec:problem_formulation}

We consider a robotic manipulation system equipped with a wrist-mounted (eye-in-hand) RGB-D camera and a parallel-jaw gripper, operating in an unknown, cluttered environment containing a set of \textit{objects} \( \mathcal{O} = \{O_1, O_2, \dots, O_m\} \). Among these, a distinct \textit{target object} \( O_{*} \) is subject to significant occlusions, often in \textit{pile} or \textit{packed}~\cite{breyer2021volumetric} configurations. The target object may be heavily or completely occluded due to \textit{coverage} — where the target is fully enclosed by surrounding objects, or due to \textit{blocking} —  where objects positioned in front obscure direct access of the sensor's line of sight.

To achieve target-oriented grasping in these scenarios, we incorporate FMs~\cite{bai2025qwen2, liu2024grounding, ravi2024sam2} for semantic scene understanding and adaptive view planning, allowing zero-shot generalization without task-specific training. In extreme cases, such as complete occlusions or no valid grasp poses available on the target, the VLM either determines an optimal grasp execution sequence from current observed objects $\mathcal{O}'$ to prioritize the removal of occluders,
 or infers the potential occluders by the VLM with VICL from historical observations. 

\section{Methods}
\label{sec:methods}

\subsection{System Overview} 

In general, VISO-Grasp integrates vision-based physical grasping and semantic scene understanding as core functional capabilities. Fig.~\ref{fig:viso} illustrates the fundamental components of our system, which comprises: 

i) Adaptive Multi-view Open-Vocabulary 3D Object Detector (AMOV3D, Sec.~\ref{sec:AMOV3D}), implemented via the VLM~\cite{bai2025qwen2}, refines 3D oriented bounding boxes (\(\mathrm{BB}\)s) and associated attributes using historical \textit{RGB-D images}, \( \{ \mathbf{I}^t, \mathbf{D}^t  \}_{t=0}^{T} \), which is aided by instance segmentations from foundation models~\cite{liu2024grounding, ravi2024sam2}. The inferred attributes from AMOV3D further inform the following active view planner.
 

ii) Target-guided View Planner (TGV-Planner, Sec. \ref{sec:TGV-Planner}) formulates a continuous velocity field \(V (\mathbf{x}, \mathcal{O}')\), which is inferred together from the current position of the camera's optical frame $\mathbf{x}$. This aims to guide the camera toward viewpoints that enhance the visibility of the target object.

iii) Real-Time Uncertainty-guided Multi-view Grasp Fusion (RT-UMGF, Sec. \ref{sec:RT-UMGF}) provides 6-DoF grasp candidates $G_t$, which are inferred from the historical \textit{depths} \( \{ \mathbf{D}^t \}_{t=0}^{T} \) by the grasp generator, which is pre-trained purely from simulated data. This aims to grasp the target object $O_*$ and remove the occluders when necessary. Below, we detail each component.


%

\subsection{Adaptive Multi-View Open-Vocabulary 3D Object Detection (AMOV3D)}
\label{sec:AMOV3D}

In general, the AMOV3D module leverages FMs to achieve robust 3D object detection and segmentation in cluttered environments.

\subsubsection{3D Scene Representation Generation}
Given an input image \( \mathbf{I} \) and a prompt-specified target object \( O_{*} \), a VLM generates structured descriptions for each detected object, prioritizing the identification of \( O_{*} \) to mitigate the risk of occlusion-induced omission. Each object instance is described using its label and three attributes: color (\( A_c \)), pattern (\( A_p \)), and spatial relation (\( A_s \)):
\begin{equation}
\mathcal{L} = \{(l_i, A_c^i, A_p^i, A_s^i) \mid i \in \mathcal{I} \},
\end{equation}
where \( \mathcal{I} \) represents the indices of detected objects, \( l_i \) denotes the instance label, and \( (A_c^i, A_p^i, A_s^i) \) are its associated attributes. This structured representation serves as a text prompt for Grounding DINO~\cite{liu2024grounding}, which predicts the corresponding 2D bounding boxes. Each bounding box is refined by SAM2~\cite{ravi2024sam2} to obtain pixel-wise instance segmentation masks. The transformation from 2D pixel coordinates \( (u, v) \) to 3D world coordinates \( (X, Y, Z) \) is performed using the depth map \( \mathbf{D} \) and the camera intrinsics \( K \)\footnote{We denote \( * \) as scalar multiplication and \( \cdot \) as the dot product.}:
\begin{equation}
(X, Y, Z)^T = \mathbf{D}(u, v) * K^{-1} \cdot (u, v, 1)^T.
\end{equation}
The 3D oriented Bounding Box (BB) is then computed using Principal Component Analysis (PCA)~\cite{dimitrov2006bounding}, aligning the box with the object's principal axes. This structured scene representation serves as the foundation for downstream planning and grasp execution.
\begin{figure*}[t]  
    \begin{minipage}{0.44\textwidth}  
        \centering
        \includegraphics[width=.95\linewidth]{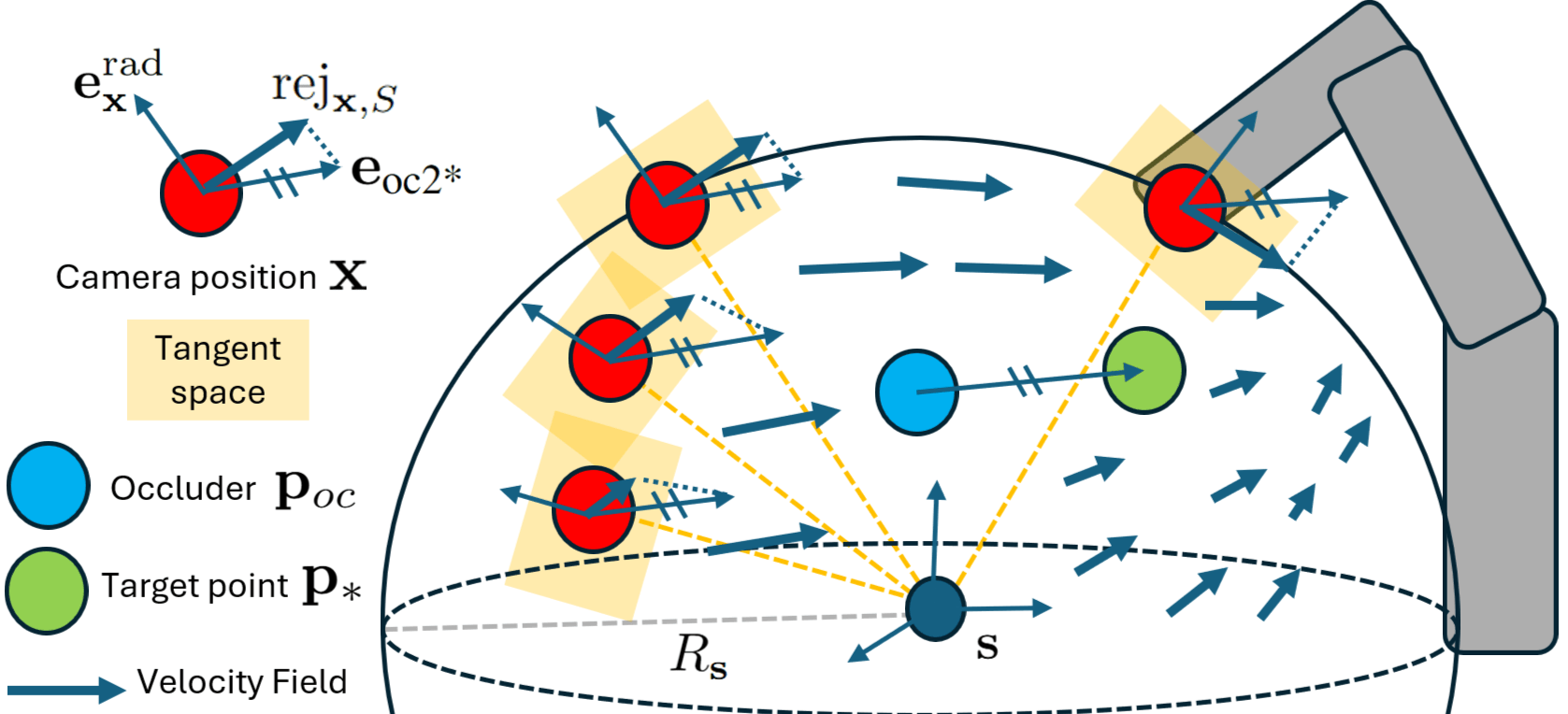}
    \end{minipage}
    \begin{minipage}{0.48\textwidth}  
        \centering
        \includegraphics[width=1.1\linewidth]{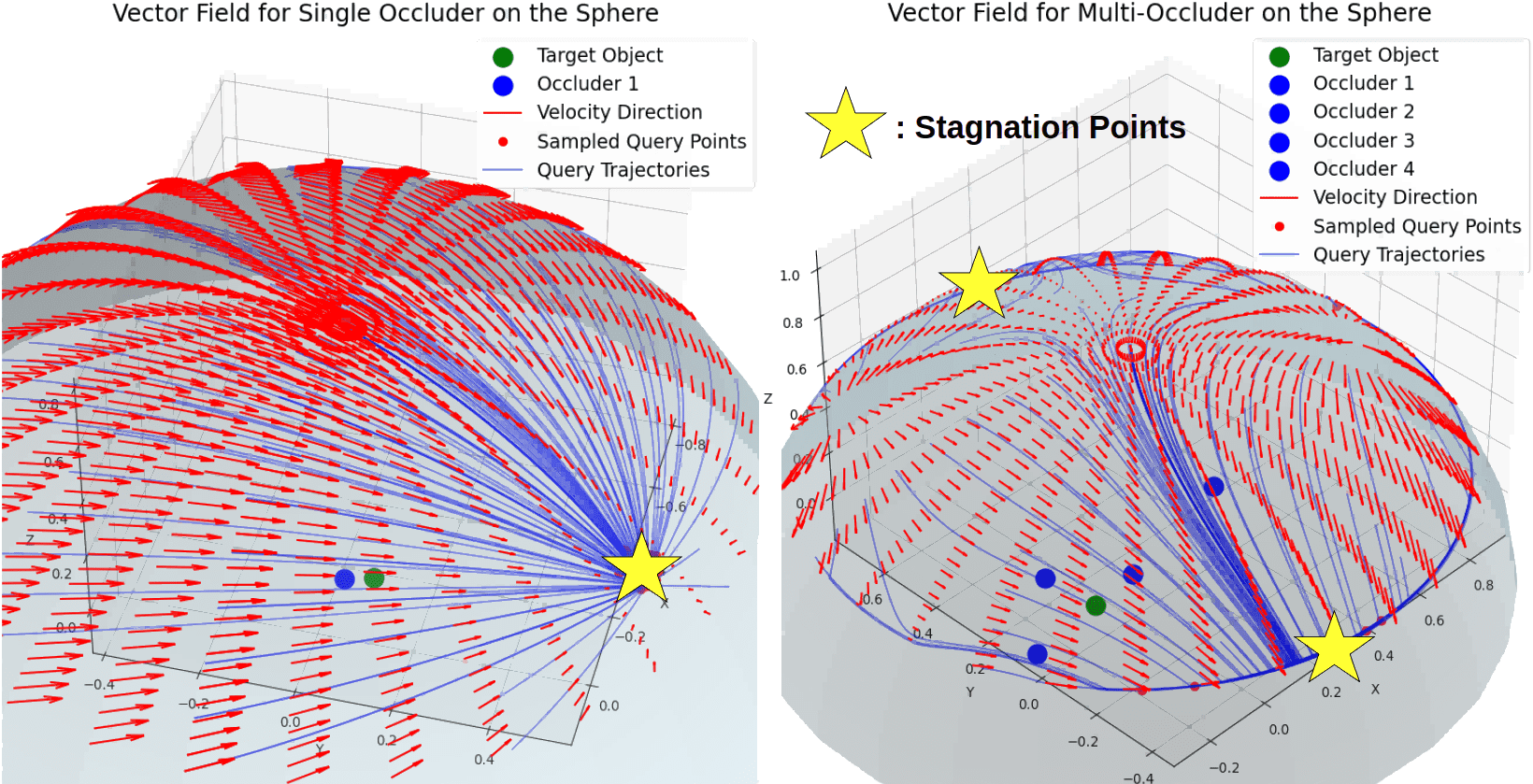}
    \end{minipage}
    \caption{Principle for TGV-Planner (left). Given single or multiple occluders (blue) and target center (green), the sampled camera viewpoints (red) follow the trajectories (blue curves) to expose the target within the spherical velocity field (red arrows) to achieve stagnation point(s) for single(middle)-/multi(right)-occluders. Note: To prevent unplannable movements, we truncate the tangential component of the field that aligns with the negative \( z \)-axis for elevation angles below \( 45^\circ \), ensuring no further downward movement.
}
\vspace{-.4cm}
    \label{nbv_field}
\end{figure*}
\subsubsection{Multi-View Object Verification and Fusion}
To provide TGV-Planner with a coherent scene representation, object detection across multiple viewpoints must maintain consistency. A historical object list \( \mathcal{O}' \), which stores all objects currently perceived within the scene, is maintained and updated at each new viewpoint to refine detected object attributes and resolve occlusions. This list persists throughout the planning process, dynamically evolving as new observations improve object descriptions and scene understanding.
Each viewpoint captured in the history \(\{\mathbf{I}^t\}_{t=0}^{T}\) provides additional observations of the scene. During Next-Best-View (NBV) planning, AMOV3D updates the detected object set \( \mathcal{O}' \) by integrating new observations \( \mathcal{L}^t \) and verifying consistency with prior data:
\begin{align}
\mathcal{O}'^{t+1} &= \mathcal{O}'^t \cup \{ O_i^t \mid O_i^t \in \mathcal{L}^t, \ \phi(O_i^t, \mathcal{O}'^t, \mathbf{I}^{t}) \}.\\
\phi &=
\begin{cases}
\text{Merge}(O_i^t, O_j^{t'}), & \text{if } \exists O_j^{t'} \in \mathcal{O}'^t
, \\
\text{Add } O_i^t, & \text{if } O_i^t \notin \mathcal{O}'^t.
\end{cases}
\end{align}
By continuously refining \( \mathcal{O}' \) and selecting viewpoints that maximize the visibility of \textit{target object} \( O_{*} \), AMOV3D improves the accuracy of object detection and occlusion reasoning, thereby enhancing the spatial reasoning accuracy of the TGV-Planner.

\subsubsection{Occlusion Reasoning and Scene Exploration}
If the target object \( O_{*} \) is not detected in the current view, the system initiates an occlusion reasoning process to infer potential obstructing objects. We employ VICL within the VLM to analyze the scene's spatial configuration and identify potential occluders based on structured object descriptions \( \mathcal{L}^t \) and the current view image \( \mathbf{I}^t \). To ensure accurate inference without additional computational cost, we enforce a Mixture-of-Reasoning-Experts (MoRE)~\cite{si2023getting} paradigm, guiding the VLM to concurrently assess spatial relations, material properties, and geometric constraints via structured prompting. The outputs are aggregated using voting to enhance occlusion identification robustness. The system then updates the grasp order, designating one potential occluder as the current grasp target. Subsequent modules adjust their reasoning and execution to remove this object, incrementally revealing the prompt-specified target object.

\subsection{Target-Guided View Planner (TGV-Planner)}
\label{sec:TGV-Planner}

The TGV-Planner performs spatial reasoning to determine whether the target object \( O_* \) can be directly grasped or if an occluder must be removed first. If occlusions prevent direct grasping, the system prioritizes occluder removal. However, when direct removal is infeasible or does not sufficiently expose \( O_* \), the \textit{Velocity-field-based NBV (V-NBV)} module optimizes the view planning. \textit{V-NBV} is applied to either enhance the visibility of \( O_* \) for grasp execution or to improve occluder removal efficiency by selecting a more informative camera perspective. To ensure robust grasp execution, the TGV-Planner operates in parallel with the grasp inference from vMF-Contact. At the same time, the grasp execution criterium evaluates the grasp confidence of \( O_* \) continuously. If this exceeds a predefined threshold, the grasp is executed immediately. Otherwise, the grasps are updated in real-time as the viewpoint is adjusted to improve visibility.

\subsubsection{Rule-Based Spatial Reasoning}
Given the BBs obtained from AMOV3D, objects in the scene are assigned spatial relationships based on predefined geometric criteria:

- \textbf{Proximity}: Two objects are in proximity if their $\mathrm{BB}$s intersect after isotropic expansion.


- \textbf{Below}: \( O_i \) is below \( O_j \) if the constraint \( \min( \mathrm{BB}_z(O_j)) - \max(\mathrm{BB}_z(O_i)) > \gamma_{\mathrm{Below}}\) is satisfied and their projections onto the \( xy \)-plane overlap. 
$\gamma_{\mathrm{Below}}$ is the height threshold.

- \textbf{High / Low}: An object \( O_i \) is considered \textbf{High} relative to \( O_j \) if 
\(
h(O_i, O_j) = \max(\mathrm{BB}_z(O_i)) - \max(\mathrm{BB}_z(O_j))
\)
satisfies
\(
h(O_i, O_j) > \gamma_{HL}\) (High) or \( h(O_i, O_j) < -\gamma_{HL}\) (Low), where \( \gamma_{HL} \) is a predefined height threshold.

The grasp execution strategy is determined by the assigned spatial relations: i) If \( O_* \) is \textbf{Below} any object, it is prioritized for grasping to expose \( O_* \);
ii) If an object in \textbf{Proximity} to \( O_* \) is also classified as \textbf{High}, the TGV-Planner evaluates whether grasping \( O_* \) is feasible or if occluder removal is required.
iii) \textbf{Low} occluders are ignored as they do not obstruct grasp execution. iv) Crucially, \textbf{High} occluders trigger the \textit{V-NBV} module to optimize viewpoint adjustment.



\subsubsection{V-NBV with Single Occluder} Given the current camera position \( \mathbf{x} \), the system queries a continuous velocity field constructed from the spatial relationships between all \textbf{High} occluders and \( O_* \) to guide viewpoint adjustment by:
\(
    V: \mathbb{R}^3 \times \mathcal{O} \to \mathbb{R}^3, \quad V(\mathbf{x}, \mathcal{O}_h) = \dot{\mathbf{x}}.
\)

We begin with single occluder scenario for a better understanding of our approach, where the principle is illustrated in Fig. \ref{nbv_field} (left). For instance, a camera is constrained on a hemisphere centered at \( \mathbf{s} \) with radius \( R_\mathbf{s} \) by position \( \mathbf{x} \). Given two center points of the target object \( \mathbf{p}_*\) and single occluder \( \mathbf{p}_{oc} \) inside the hemisphere respectively, the objective is to generate camera motions that naturally follow the geodesic that improves the visibility of the target. To achieve this, the velocity at \( \mathbf{x} \) is computed as follows: 

First, the normalized vector \( \mathbf{e}_{\text{oc2*}} \) from \( \mathbf{p}_{oc} \) to \( \mathbf{p}_*\) and the radial vector from the sphere center \( \mathbf{s} \) to \( \mathbf{x} \) is given by:
\begin{equation}
\mathbf{e}_{\text{oc2*}} = \frac{\mathbf{p}_*- \mathbf{p}_{oc}}{\| \mathbf{p}_*- \mathbf{p}_{oc} \|}, \quad \mathbf{e}_{\mathbf{x}}^{\mathrm{rad}} = \frac{\mathbf{x} - \mathbf{s}}{R_\mathbf{s}}.
\end{equation}
To obtain the velocity direction $\mathbf{e}_{\dot{\mathbf{x}}}$ within the tangent space of sphere $(\mathbf{s}, R_\mathbf{s})$ allocated on 
$\mathbf{x}$, which needs to align with \( \mathbf{e}_{\text{oc2*}} \), we further calculate the normalized rejection of \( \mathbf{e}_{\text{oc2*}} \) onto the radial vector $\mathbf{e}_{\mathbf{x}}^{\mathrm{rad}}$, namely $\mathbf{e}_{\dot{\mathbf{x}}}$ by:
\begin{equation}
    \mathrm{rej}_{\mathbf{x},S}= \mathbf{e}_{\text{oc2*}} - (\mathbf{e}_{\text{oc2*}} \cdot \mathbf{e}_{\mathbf{x}}^{\mathrm{rad}}) *\mathbf{e}_{\mathbf{x}}^{\mathrm{rad}}, 
    \quad 
    \mathbf{e}_{\dot{\mathbf{x}}} = \frac{\mathrm{rej}_{\mathbf{x},S}}{\|\mathrm{rej}_{\mathbf{x},S}\|}.
\end{equation}
Notably, we define the field strength coefficient \( \beta \), which is computed based on the angular relationship between \( \mathbf{x} \), \( \mathbf{p}_{oc} \), and \( \mathbf{p}_*\) using:
\begin{equation}
    \beta = \frac{1}{\pi} * \arccos\Big(\frac{\mathbf{e}_{\text{oc2*}} \cdot (\mathbf{x} - \mathbf{p}_{oc})}{\| \mathbf{x} - \mathbf{p}_{oc} \|}\Big).
\end{equation}

This ensures that the field's strength 
\(\beta\) smoothly decreases from 1 to 0 as the camera moves from the occluder's side to the opposite of the target, converging at the stagnation point (\(\beta=0\)), which represents the final Next-Best View (NBV). Finally, the velocity field is calculated by:
\(\dot{\mathbf{x}}=V(\mathbf{x}, \mathcal{O}_h) =\beta \cdot \mathbf{e}_{\dot{\mathbf{x}}}.
\)
\subsubsection{Field Superposition for Multi-occluders} For multiple occluders, the camera motion is determined by the weighted combination of individual occluder's influence using the superposition principle. Given a set of $M$ occluders, the aggregated velocity is computed as the weighted sum  by $\beta$ following: \(\dot{\mathbf{x}}_M = \sum_{i=1}^{M} \beta_m \cdot \mathbf{e}_{\dot{\mathbf{x}},m}\). 

Here we illustrate two examples of single/multi-occluders in Fig. \ref{nbv_field} (middle \& right). In addition, to ensure stable and controlled motion, \textit{both the center and the bounding box corners} of each potential occluder are incorporated as occluder points $\mathbf{p}_{oc}$s into the field generation.

\subsection{Real-Time Uncertainty-guided Multi-view Grasp Fusion (RT-UMGF)} 

\begin{figure}[]
    \includegraphics[width=.45\textwidth]{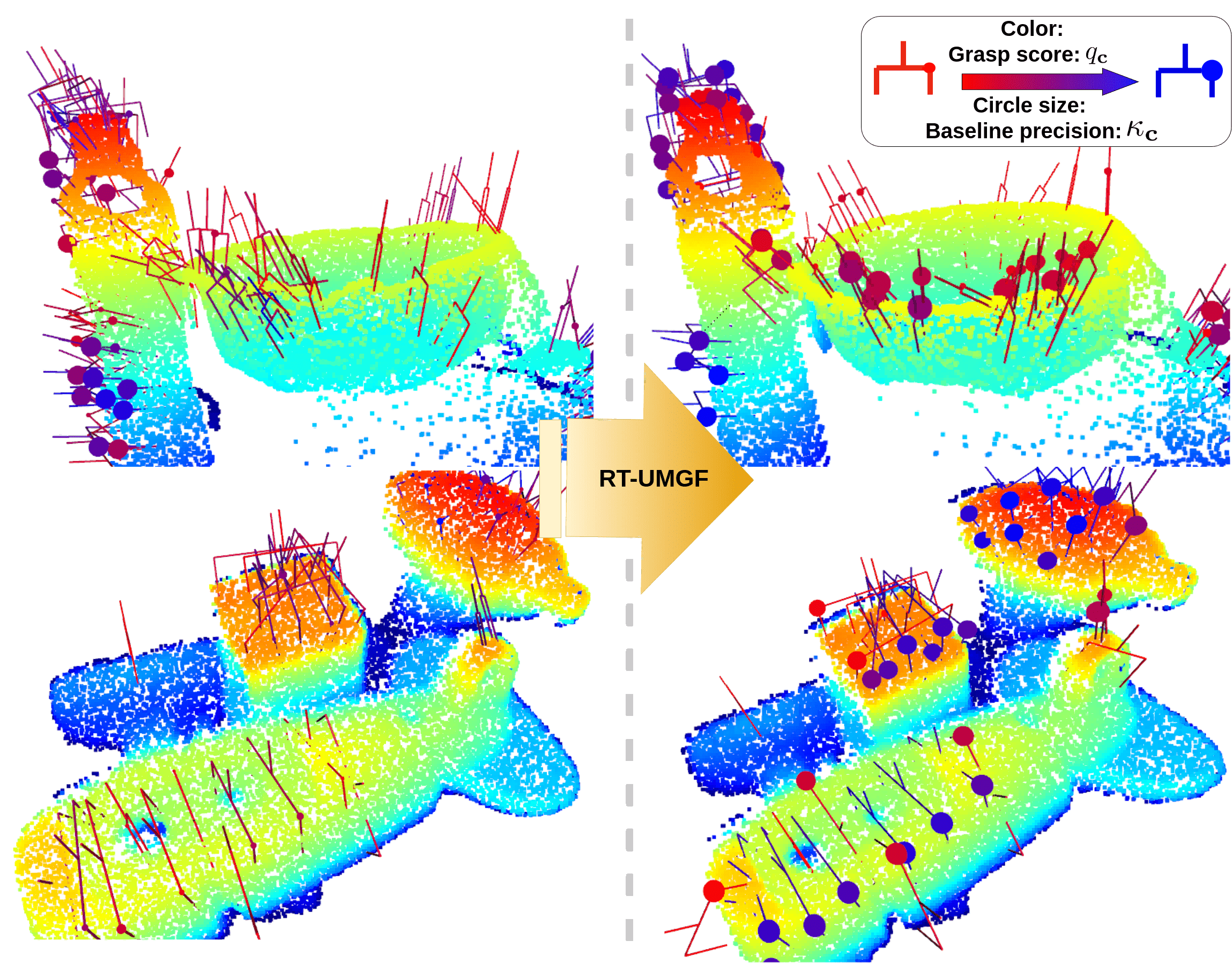}
    \caption{RT-UMGF update, where grasps are merged according to corresponding categories. Confident grasps on high grasp quality $q_\mathbf{c}$ and directional precision $\kappa_\mathbf{c}$ can be observed consequently and fused to enhance their confidence further.} 
    \label{umgf}
    \vspace{-.4cm}
\end{figure}

\label{sec:RT-UMGF}
In general, RT-UMGF continuously refines grasp predictions by fusing multi-view observations. It applies Bayesian updates using a von Mises-Fisher (vMF) distribution, ensuring stable grasp selection in cluttered environments (Fig. \ref{umgf}). To maximize the utility of multi-view sensor data and enhance grasp generation robustness, we employ the vMF-Contact~\cite{shi2024vmf}, which is adapted to real-time (10Hz) inference through pose-centric uncertainty modeling and buffered online Bayesian fusion. 
Our approach diverges from recent work on real-time grasp fusion~\cite{zhang2024gamma}, which rely solely on grasp quality scores and lack geometric grasp representation~\cite{sundermeyer2021contact}, by integrating directional von Mises-Fisher (vMF) distribution into our Bayesian update process~\cite{mardia1976bayesian}.

In runtime, assume that the raw point cloud is captured in the time frame \(t\), a set of contact grasps \cite{sundermeyer2021contact} is inferred as \(G_t = \{\mathbf{g}^t_n = (\mathbf{c}, \mathbf{\mu}_\mathbf{c}, \mathcal{\kappa}_\mathbf{c}, \Delta_\mathbf{c}, w_{\mathbf{c}}, q_\mathbf{c})^t_n\;|\;n\in N\}\) that parameterizes: 1) The queried contact points \(\mathbf{c}\in \mathbf{R}^3\); 2) Baseline vector distributions: \(p(\mathbf{b}|\mathbf{c}) = \mathrm{vMF}(\mathbf{b}|  \mathbf{\mu_\mathbf{c}}, \kappa_\mathbf{c}) =  Z(\kappa_\mathbf{c}) \exp(\kappa_\mathbf{c}  \mathbf{\mu_\mathbf{c}}^\top \mathbf{b}) \), with $\mu_c$ the mean direction of the baseline and $\kappa_c$ as the directional precision. \(Z(\kappa_\mathbf{c})\) is the normalization factor; 3) The quantized approach vector, represented by \(\Delta_\mathbf{c}=\{\delta^k_\mathbf{c}\}_{k=0,...,K}\) represents discrete categorical bin scores that define a direction constrained to lie on a plane perpendicular to a given baseline. 4) The grasp width \(w_{\mathbf{c}}\) and 5) The contact quality score $q_\mathbf{c}$.

\subsubsection{Grasp Categorization} Technically, newly generated grasps \(G'_t\) are categorized into two distinct groups based on their similarity to previously fused grasps \(G_{t-1}\): The first contact point set $C'^{\text{new}}_t$ comprises grasps located significantly away from any existing fused grasps, which then undergo a process of \textit{self-fusion} before being directly involved into the grasp buffer. This typically includes grasps associated with previously unseen objects. Conversely, if the new grasps are proximal to any previously fused grasps, they are fused with the closest existing grasp by \textit{cross-fusion}, denoted as $C'^{\text{pro}}_t$. As inspired by~\cite{zhang2024gamma}, the rule of distinction is:
\begin{align}
    C'^{\text{pro}}_t =  \big\{\ \mathbf{c}_j^t \in C'_t | \;
     &\|\mathbf{c}_j^t - \mathbf{c}_i^{t-1}\|_2 < \gamma_d, \\ 
    & 1-\cos(\mu_{\mathbf{c}_i}^{t-1}, \mu_{\mathbf{c}_j^t}) > \gamma_\theta\big\}, \\
    C'^{\text{new}}_t &= C'_t \setminus C'^{\text{pro}}_t .
\end{align}
To ensure the multi-modality of scene-level grasps, the similarity criterion between proximal and new grasps is determined based on 1) contact point distance limit $\gamma_d$ and 2) cosine similarity of baseline mean directions constraint $\gamma_\theta$. Notably, different from \textit{cross-fusion}, which applies these criteria based on existing fused grasp $\mathbf{g}^{t-1}_i$, \textit{self-fusion} does not rely on a predefined cluster center, where we applied DBSCAN~\cite{ester1996density} to identify the clusters.

\subsubsection{Contact-grasp Fusion} Both \textit{self/\textit{cross-fusion}} processes are designed as follows: For contact point positions \(\mathbf{c}^t_j\), we adopt the same regime as~\cite{zhang2024gamma} utilizing weighted sum by grasp quality $q$: \(\mathbf{c}^t_i = \frac{q^{t-1}_i\mathbf{c}^{t-1}_i + \sum_{j}(q_j\mathbf{c}^{t}_j)}  {q^{t-1}_i + \sum_{j}(q_j)}\) with grouped element indices $j\in J$. For baselines and approach vectors, the conjugate prior of vMF baseline distribution is initialized as: \(\mathbf{\mu_\mathbf{c}} \sim \mathrm{vMF}( \mu^{0}_{\mathbf{c}}, \kappa^0_{\mathbf{c}})\) for $t=0$. 

For the Bayesian inference in time frame \(t\), we may update the posterior distribution following the rule of the exponential family~\cite{charpentier2021natural} by:
\begin{equation}
\label{postup}
    \mu^{t}_{\mathbf{c}} = \frac{\kappa^{t-1}_\mathbf{c} \mu^{t-1}_{\mathbf{c}} + \sum_{j} \kappa^{t}_{\mathbf{c}_j}  \mu^{t}_{\mathbf{c}_j}}{\kappa^{t-1}_\mathbf{c} + \sum_{j} \kappa^{t}_{\mathbf{c}_j}},  \kappa^{t}_\mathbf{c} = \kappa^{t-1}_\mathbf{c} + \sum_{j} \kappa^{t}_{\mathbf{c}_j}.
\end{equation}
The approach categories are updated by:
\begin{equation}
        \delta^{t, i}_\mathbf{c} = \delta^{t-1, i}_{\mathbf{c}} + \sum_{j}\delta^{t, i}_{\mathbf{c}_j}.
        \label{postup2}
\end{equation}
Here \(\kappa^t_\mathbf{c}\) represents the precision on the observed mean likelihood \(\mu_\mathbf{c}\). We refer interested readers for the theoretical background to~\cite{shi2024vmf}. 

Throughout the runtime, we define three termination criteria for executing a grasp:  
i) The highest estimated grasp quality \( q^*_c \) must exceed a predefined threshold \( q_{\max} \). 
ii) The directional uncertainty of the corresponding grasp from condition i), \( \kappa^*_c \), must surpass a given threshold \( \kappa_{\max} \).  
iii) When the magnitude of the queried velocity field approaches a stagnation point, i.e., \( \|\dot{\mathbf{x}}\| \approx 0 \), the current best grasp is executed immediately if condition i) is fulfilled. Otherwise the grasping prioritization, where the optimal grasp order of potential obstructers will be queried from the VLM and the target will be switched.

\section{Experiments}
During the experiments, we aim to analyze the contributions of each component in VISO-Grasp. We investigate the advantages of TGV-Planner over fixed initial (\textit{Init view}) and \textit{Top-down} view setups in enhancing the semantic understanding of the VLM, improving the grasp update process, and refining grasp proposals. Furthermore, we include a baseline of identical settings as VISO-Grasp without grasp fusion (\textit{w/o GF}), which aims to assess the effectiveness of RT-UMGF as another focus of our studies. In this setting, grasp predictions are inferred directly on the NBV in the experiments. Additionally, we incorporate reconstruction-based NBV planning approaches~\cite{breyer2022closed} (\textit{Breyer's}) for comparison.

\subsection{Experiment setups}
Our experimental setup consists of a UR10e robotic arm integrated with an Orbbec Femto Mega RGB-D camera for scene perception and a Robotiq 2F85 parallel-jaw gripper for object manipulation (See Fig. \ref{fig:highlight}). In AMOV3D, we employ Qwen2.5-VL-72B-Instruct-AWQ~\cite{bai2025qwen2} as the Visual-Language Model (VLM) to enhance high-level reasoning and scene understanding. For RT-UMGF, we leverage the vMF-Contact model with PointNeXt-B~\cite{qian2022pointnext} as the backbone, which is pre-trained on a purely simulated dataset according to~\cite{shi2024vmf}, to infer the scene-level grasp and uncertainty prediction. 
\begin{table}
  \renewcommand\arraystretch{.92} 
  \vspace{.05cm}
  \caption{Real-world experiment setups and results.}
  \label{tab:real-world-exp}
  \setlength{\tabcolsep}{.2pt} 
  \begin{tabular}{c c c c c c}
        \toprule
        \multirow{2}{*}{\textbf{Setup}} & \multirow{2}{*}{\textbf{Initial View}}&\multicolumn{4}{c}{Scene index: \textbf{$O_*$}} \\ 
        \cmidrule(lr){3-6}
      \multicolumn{2}{c}{} & \textbf{Method} & \textbf{FS(/5)} & \textbf{ \#GA} & \textbf{GSR(\%)} \\
      \midrule
        \multirow{6}{*}{\centering
        \includegraphics[width=.225\columnwidth]
        {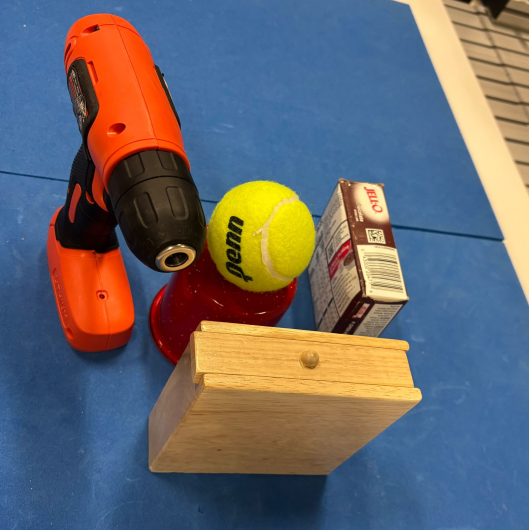}
        }
        & \multirow{6}{*}{\centering
        \includegraphics[width=.225\columnwidth]
        {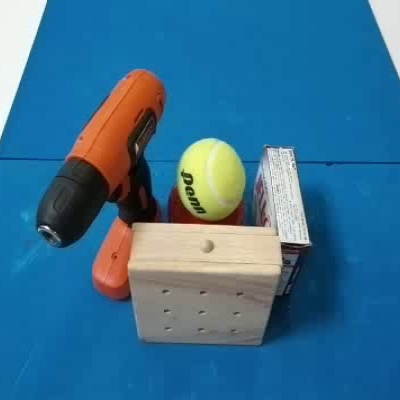}
                }
        &\multicolumn{4}{c}{(1): \textbf{Red Cup}} \\
        \cmidrule(lr){3-6}
        &&\textit{Breyer's} &2&4.0& 27.8\\
        & & \textit{Top-down} &2&3.6&61.1\\ 
        & & \textit{Init view} &\textbf{5}&\textbf{3.4}&88.2\\
        & & \textit{w/o GF} &\textbf{5}&3.6&\textbf{94.44}\\
        & & \textit{Our's} &\textbf{5}&\textbf{3.4}&88.2\\
        \midrule
        \multirow{6}{*}{\centering
        \includegraphics[width=.225\columnwidth]
        {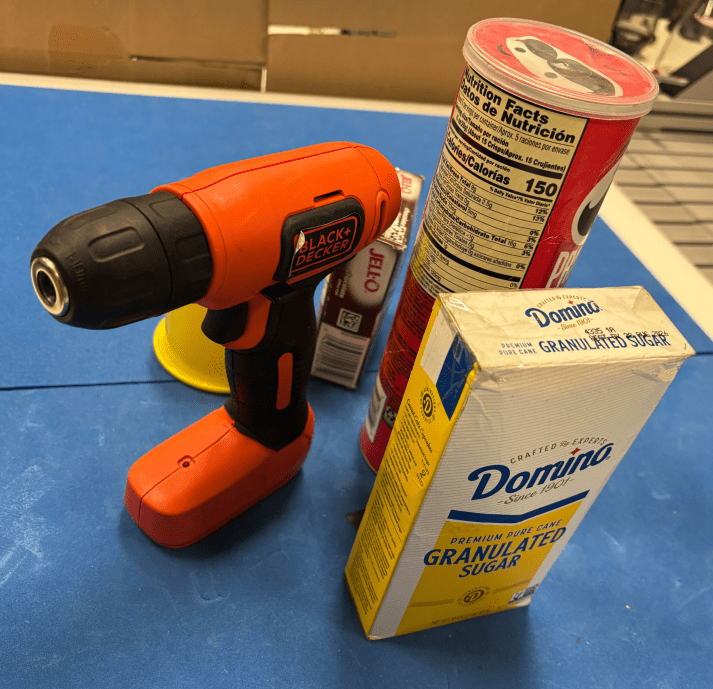}
        }  
        & \multirow{6}{*}{\centering
        \includegraphics[width=.225\columnwidth]
        {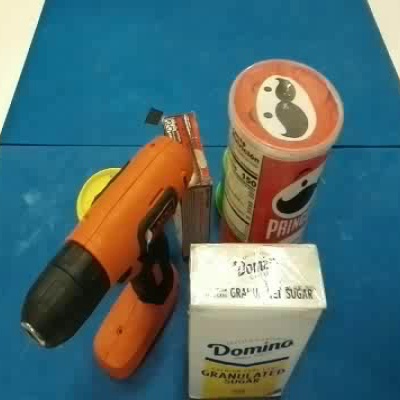}
        }
        &\multicolumn{4}{c}{(2): \textbf{Jello Box}} \\
        \cmidrule(lr){3-6}
        && \textit{Breyer's} &0&4.0&6.3\\
        & & \textit{Top-down} &\textbf{4}&\textbf{2.2}&70.0\\
        & & \textit{Init view} &3&3.0&33.3\\
        & & \textit{w/o GF} &3&3.6&66.7\\
        & & \textit{Our's} &\textbf{4}&3.0&\textbf{80.0}\\
        \midrule
        \multirow{6}{*}{\centering\includegraphics[width=.225\columnwidth]
        {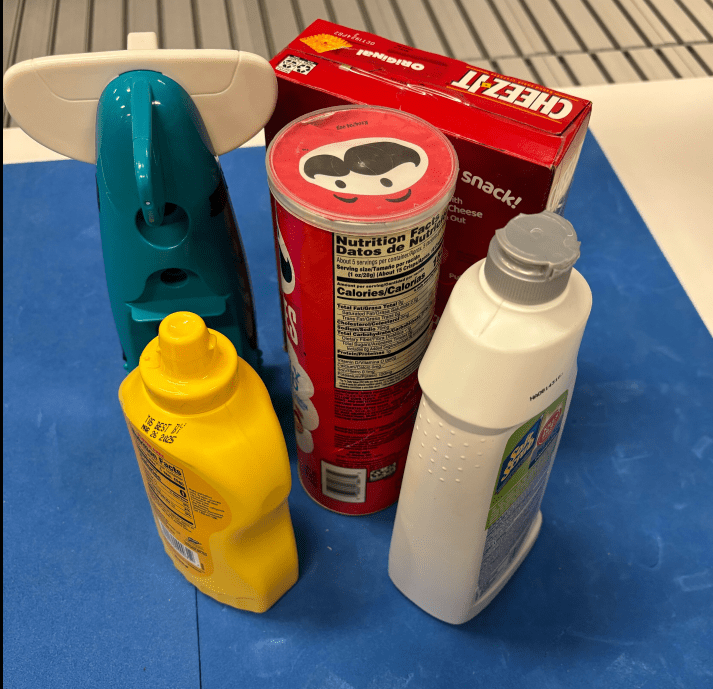}
        } 
        & \multirow{6}{*}{\centering
        \includegraphics[width=.225\columnwidth]
        {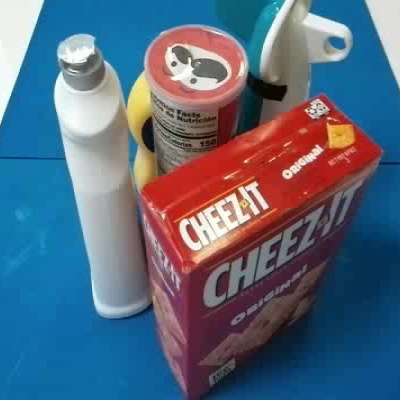}
        }
        &\multicolumn{4}{c}{(3): \textbf{Pringles}} \\
        \cmidrule(lr){3-6}
        && \textit{Breyer's} &0&4.0&11.8\\
        & & \textit{Top-down} &2&2.8&30.8\\
        & & \textit{Init view} &4&3.2&68.8\\
        & & \textit{w/o GF} &\textbf{5}&2.0&\textbf{90.0}\\
        & & \textit{Our's} &\textbf{5}&\textbf{1.6}&87.5\\
        \midrule
        \multirow{6}{*}{\centering
        \includegraphics[width=.225\columnwidth]
        {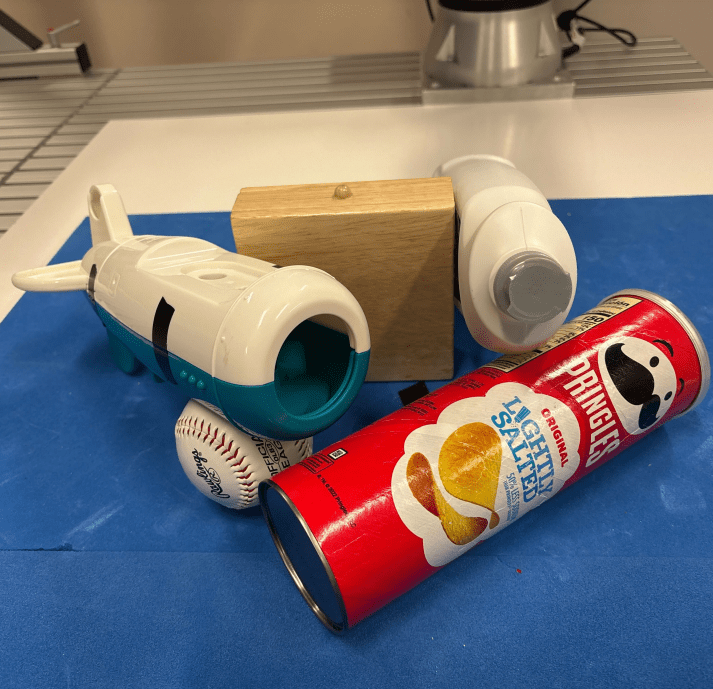}
        }
        & \multirow{6}{*}{\centering\includegraphics[width=.225\columnwidth]
        {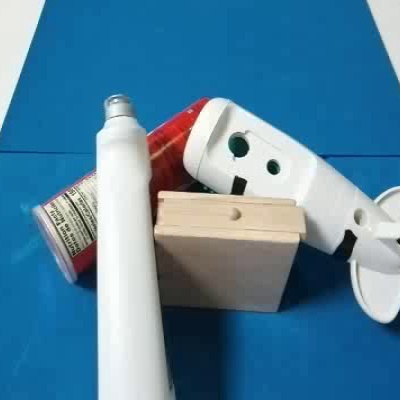}}
        &\multicolumn{4}{c}{(4): \textbf{Baseball}} \\
        \cmidrule(lr){3-6}
        && \textit{Breyer's} &1&4.0&18.8\\
        & & \textit{Top-down} &4&3.2&62.5\\
        & & \textit{Init view} &2&3.2&62.5\\
        & & \textit{w/o GF} &3&3.4&82.4\\
        & & \textit{Our's} &\textbf{5}&\textbf{2.8}&\textbf{85.7}\\
        \midrule
        \multirow{6}{*}{\centering
        \includegraphics[width=.225\columnwidth]
        {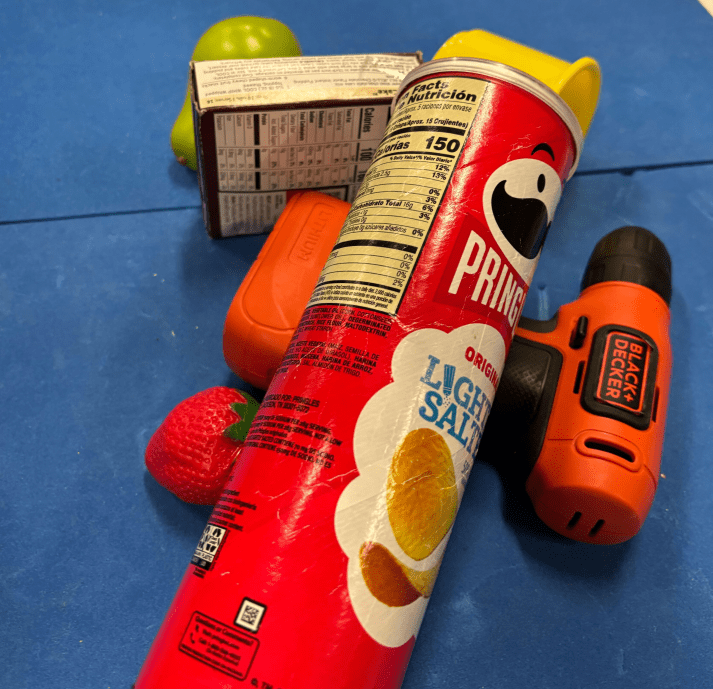}
        } & 
        \multirow{6}{*}{\centering
        \includegraphics[width=.225\columnwidth]
        {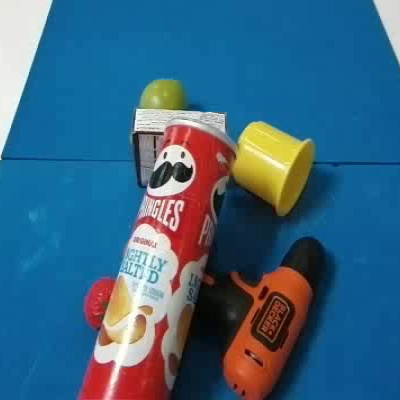}
        }
        &\multicolumn{4}{c}{(5): \textbf{Green Pear}} \\
        \cmidrule(lr){3-6}
        && \textit{Breyer's} &0.0 &4.0&0.0\\
        & & \textit{Top-down} &4&2.6&46.1\\
        & & \textit{Init view} &3&3.6&88.9\\
        & & \textit{w/o GF} &\textbf{5}&3.2&68.8\\
        & & \textit{Our's} &\textbf{5}&\textbf{2.4}&\textbf{100.0}\\
        \midrule
        \multirow{6}{*}{\centering
        \includegraphics[width=.225\columnwidth]
        {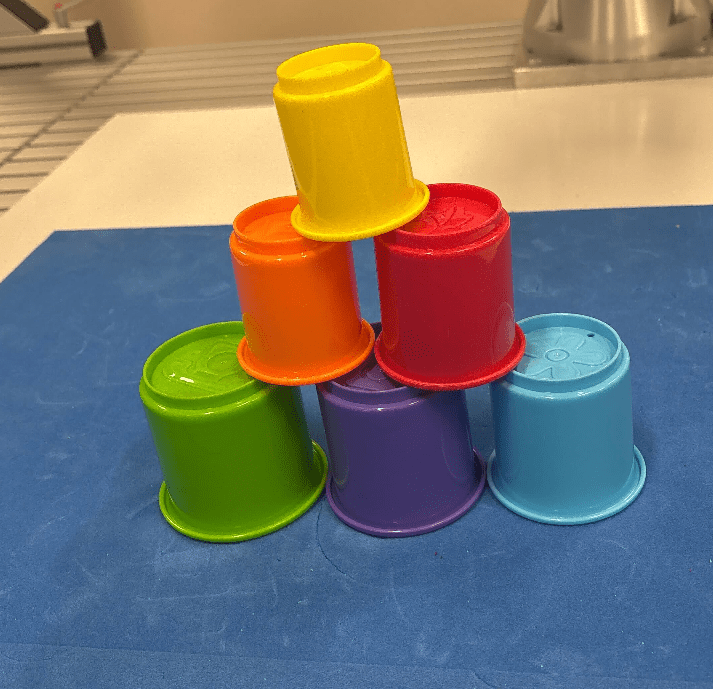}
        } 
        & \multirow{6}{*}{\centering
        \includegraphics[width=.225\columnwidth]
        {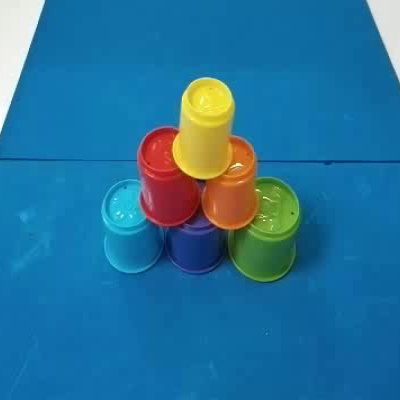}
        }
        &\multicolumn{4}{c}{(6): \textbf{Purple Cup}} \\
        \cmidrule(lr){3-6}
        && \textit{Breyer's} &2&4.0&\textbf{85.0}\\
        & & \textit{Top-down} &1&4.0&70.0\\
        & & \textit{Init view} &1&\textbf{3.6}&27.8\\
        & & \textit{w/o GF} &1&4.0&45.0\\
        & & \textit{Our's} &\textbf{3}&4.0&\textbf{85.0}\\
        \midrule
        \multirow{6}{*}{\centering
        \includegraphics[width=.225\columnwidth]
        {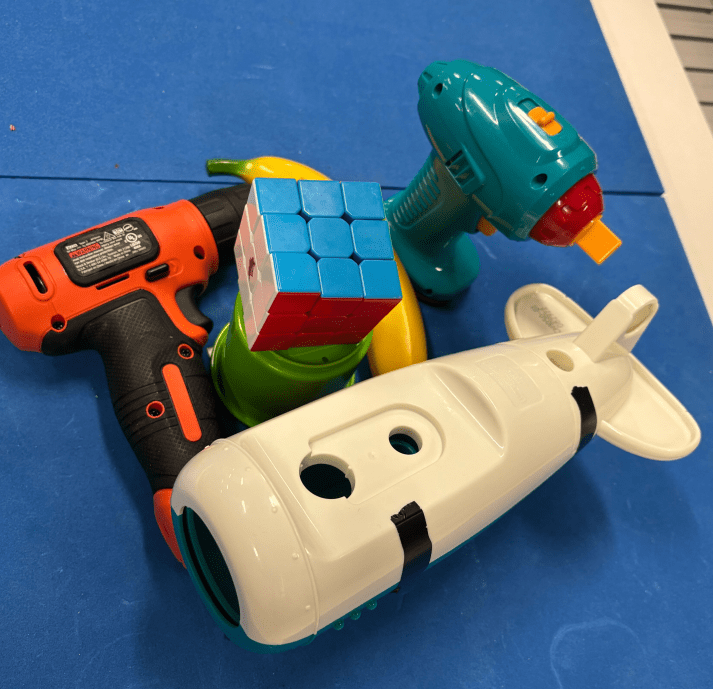}        
        }  & 
        \multirow{6}{*}{\centering
        \includegraphics[width=.225\columnwidth]
        {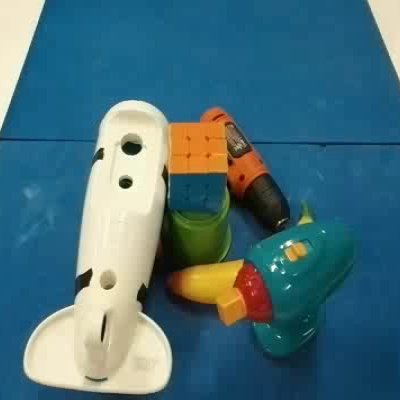}
        }
        &\multicolumn{4}{c}{(7): \textbf{Yellow Peach in the Green Cup}} \\
        \cmidrule(lr){3-6}
        && \textit{Breyer's} &0&4.0&22.2\\
        & & \textit{Top-down} &2&4.0&75.0\\
        & & \textit{Init view} &3&3.8&63.2\\
        & & \textit{w/o GF} &\textbf{4}&3.6&\textbf{83.3}\\
        & & \textit{Our's} &\textbf{4}&\textbf{3.2}&75.0\\
        \midrule
        \multirow{6}{*}{\centering
        \includegraphics[width=.225\columnwidth]
        {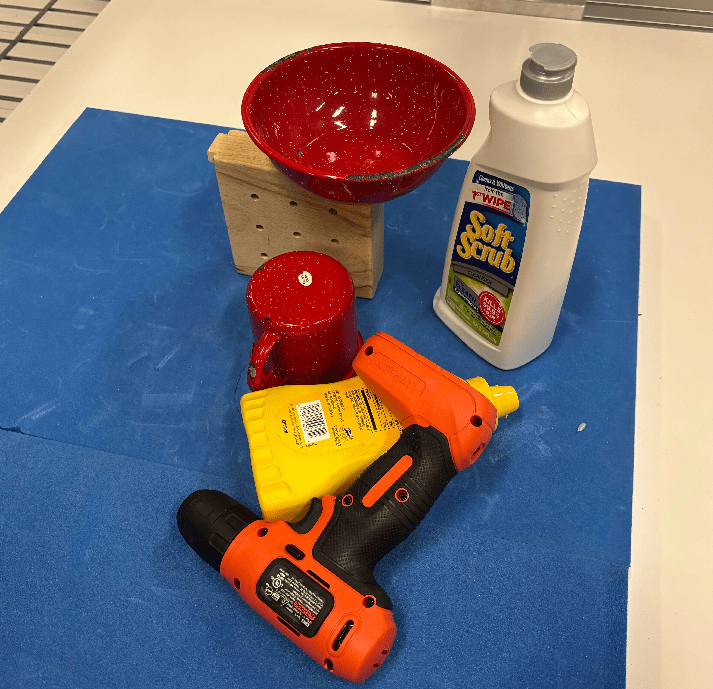}
        } & 
        \multirow{6}{*}{\centering
        \includegraphics[width=.225\columnwidth]
        {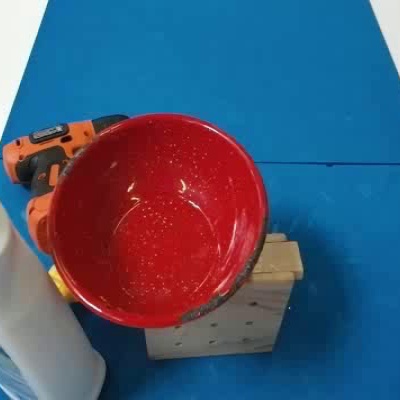}
        }
        &\multicolumn{4}{c}{(8): \textbf{Tennis Ball in the Red Cup}} \\
        \cmidrule(lr){3-6}
        && \textit{Breyer's} &0&4.0&21.1\\
        & & \textit{Top-down} &2&4.0&45.0\\
        & & \textit{Init view} &3&4.0&80.0\\
        & & \textit{w/o GF} &2&\textbf{3.8}&55.0\\
        
        & & \textit{Our's} &\textbf{4}&\textbf{3.8}&\textbf{84.2}\\
        \bottomrule
    \end{tabular}
\end{table}
The following setups are considered in our real-world experiments for ablation studies: \textit{Top-down} and \textit{Init view} are used to evaluate the performance w/o \textit{V-NBV} in the TGV-Planner. In these experiments, all other components, including AMOV-3D, spatial reasoning in the TGV-Planner and RT-UMGF, remain identical as with our approach (denoted as \textit{Our's}). Specifically, the \textit{Init view} setup maintains a fixed viewpoint at a $45^\circ$ elevation angle in the horizontal coordinate, identical to the configuration in \textit{Ours}. Besides, to evaluate RT-UMGF, we include a comparison setting that excludes the grasp fusion process ( \textit{w/o GF}), where the best grasp is determined by the current inference after reaching the NBV. Finally, we incorporate \textit{Breyer's} closed-loop NBV planner~\cite{breyer2022closed} as a baseline. For fair comparisons, we manually annotated the \(\mathrm{BB}\)s 
involving all objects in proximity to the targets. These \(\mathrm{BB}\)s are axis-aligned with the robot frame, allowing the planner to perform targeted view planning. 

To systematically evaluate the performance of VISO-Grasp, we conducted comprehensive experiments across eight distinct scenes, each varying in difficulty, occlusion severity, and occlusion type (Table. \ref{tab:real-world-exp}). The following metrics are used in our experiments: Final Success \textbf{(FS)} quantifies the number of successful grasps of the target object; Grasp Attempts \textbf{(\#GA)} counts the average number of grasp attempts required for each scene until the final target object is successfully grasped. If the target is not grasped by the end of the trial, this value is not recorded; Grasp Success Rate \textbf{(GSR)} measures the overall success rate of grasping any objects throughout the entire trial.   A trial is aborted if any of the following conditions occur four times after the scene is reset to its previous state:  i) Grasp failure; ii) More than one object is affected after grasp execution, leading to unintended scene changes or exposure of the target object; iii) A collision of the gripper fingers triggers an emergency stop of the robot. 

\begin{table}[h]
    \centering
    \caption{Average performance results}
    \label{tab:average}
    \setlength{\tabcolsep}{14pt}
    \begin{tabular}{cccc}
        \toprule
        \textbf{Method} & \textbf{AFSR} (\%) & \textbf{\#AGA} & \textbf{AGSR} (\%) \\
        \midrule
        \textit{Breyer’s}    & 12.50  & 4.00  & 22.5   \\
        \textit{Top-down}     & 52.50  & 3.30 & 56.50   \\
        \textit{Init view}   & 60.00 & 3.48  & 64.08  \\
        \textit{w/o GF}     & 70.00  & 3.43& 73.19      \\
        \textit{Ours}        & \textbf{87.50} & \textbf{3.10} & \textbf{83.86}  \\
        \bottomrule
    \end{tabular}
    \vspace{-.4cm}
\end{table}

\subsection{Experiment evaluation}
The experimental results from $8$ trials across various scenes are presented in Table.~\ref{tab:real-world-exp}, which demonstrate the effectiveness of VISO-Grasp (\textit{Ours}), achieving high success in reaching the target with minimal grasp attempts across diverse occlusion scenarios. Compared to baselines with static viewpoints (\textit{Top-down} and \textit{Init view}), VISO-Grasp leverages the synergy between TGV-Planner and RT-UMGF to actively explore viewpoints and refine grasp selection. This facilitate the grasp success on highly occluded or fully unobservable targets. For instance, in Scene (1), where the target object is partially obstructed by a wooden block and positioned beneath a tennis ball, both \textit{Top-down} and \textit{Init view} exhibit limited success due to frequent collisions with the wooden block. In Scene (3), although \textit{Top-down} provides an unobstructed view, it frequently misclassifies objects due to insufficient semantic understanding by the VLM (e.g., recognizing the "Pringle box" as a "Red can"). In contrast, VISO-Grasp mitigates these limitations by leveraging multi-view integration from AMOV3D, refining predictions through diverse observations facilitated by the TGV Planner. As a result, it achieves success in all five trials with the GSR exceeding $80\%$. In addition, Breyer's NBV planner, which lacks object-centric view planning, struggles to effectively expose the target object, even when predefined \(\mathrm{BB}\)s are provided around the target. This highlighted the importance of diverse viewpoints generated by TGV-Planner, which enhance comprehensive scene understanding and facilitate more informed grasp planning in general.

In addition, a detailed ablation study on RT-UMGF reveals its crucial role in reducing grasp attempts and improving grasp success by prioritizing high-quality, low-uncertainty grasps over time. In Scene (4), the setting without grasp fusion (w/o GF) achieves a grasp success rate (GSR) of $82.4\%$ with an average of $3.2$ grasp attempts. By contrast, VISO-Grasp further increases GSR to $85.7\%$ while reducing grasp attempts to $2.8$,  confirming that multi-view grasp aggregation refines grasp selection by integrating uncertain and low-quality grasp hypotheses, thereby enhancing selection robustness through the fusion of multiple grasp candidates.


The importance of NBV and grasp fusion becomes more evident when analyzing scenarios where information is severely lacking and semantic information is easily misinterpreted. In Scene (6), where the purple cup is at the bottom of the scene, understanding spatial hierarchy is crucial to successfully grasp the target, as it is one of the most challenging scenes overall. \textit{Breyer's} NBV planner successfully removes cups on top with high GSR but occasionally fails to grasp purple cup due to insufficient object-centric guidance, while \textit{Ours} achieves comparable success with identical GSR. 
The synergy between NBV and grasp fusion is crucial to the success of \textit{Ours}, since NBV ensures that occluded targets become visible, but grasp predictions from new viewpoints may still exhibit uncertainty.

In general, Table. \ref{tab:average} presents the overall performance of each method in terms of Average Final Success Rate \textbf{(AFSR)}, Average Grasp Attempts \textbf{(\#AGA)} and Average Grasp Success Rate \textbf{(AGSR)}. \textit{Breyer's} reflects its limits in revealing the target’s critical geometry and underscores how the lack of target-focused visibility. \textit{Top-down} and \textit{Init view} achieve moderate performance, which struggles in heavy occlusion and fixed view limit. \textit{Top-down} benefits from its view for some cases and tries to grasp the target in an efficient manner, but results in lower AGSR due to the lack of semantic information. \textit{w/o GF} outperforms all static baselines with $70.00\%$ of AFSR by employing next-best-view planning to reduce occlusion. However, without RT-UMGF, it still produces suboptimal grasps more frequently. 
In summary, VISO-Grasp (\textit{Ours}) demonstrates the best overall performance by $87.50\%$ combining active NBV exploration with real-time grasp fusion. 


\section{Conclusion}
\label{sec:conclusion}

We develop VISO-Grasp, a novel vision-language-informed system for target-oriented grasping in highly unstructured environments including entire invisibility. By integrating a Vision-Language Model (VLM) with object-centric View planning and real-time uncertainty-driven grasp fusion, our system enhances scene understanding and improves grasp success through continuous velocity fields and semantic spatial reasoning for adaptive grasping in occluded environments with complete invisibility. VISO-Grasp leverages robust multi-view aggregation to refine grasp selection by integrating uncertain grasp hypotheses, ensuring superior stability and accuracy. Experimental results show that VISO-Grasp achieves a final success rate by $87.5\%$ while requiring the fewest grasp attempts among all baselines, demonstrating its efficiency in occlusion-aware grasping.

Certain limitations remain for VISO-Grasp that warrant further investigation. First, the system's reliance on the VLM introduces a computational bottleneck, as VLM inference incurs non-negligible latency, especially for VICL's context, which demands extensive reasoning over multi-modal inputs. Second, our experiments assumes a quasi-static scene, which does not account for highly dynamic objects or external disturbances. Future work could explore adaptive NBV strategies in real-time scene variations.


\bibliographystyle{IEEEtran}
\bibliography{bibfile}



\end{document}